\documentclass[10pt]{article}

\usepackage{acl}

\usepackage[T1]{fontenc}
\usepackage[utf8]{inputenc}
\usepackage{times}
\usepackage{microtype}
\usepackage{latexsym}
\usepackage{amsmath}
\usepackage{amssymb}
\usepackage[most]{tcolorbox}
\usepackage{booktabs}
\usepackage{array}
\usepackage{graphicx}
\usepackage{xcolor}
\usepackage{pgfplots}
\usepgfplotslibrary{groupplots}
\pgfplotsset{compat=1.18}
\usepackage{placeins}
\usepackage{tikz}
\usetikzlibrary{arrows.meta,positioning,fit,calc,backgrounds,shapes.geometric}
\hypersetup{colorlinks=true,citecolor=blue,linkcolor=blue,urlcolor=blue}
\newcommand{\best}[1]{\textbf{#1}}
\newcommand{\sys}{\textsc{LLM-Instruct}}

\newcounter{modelbox}[subsubsection]
\renewcommand{\themodelbox}{\thesubsubsection}

\title{\sys{} at UZH Shared Task 2026:\\Constraint-Aware Retrieval and Selective Debate for \\Paragraph-Level Argument Mining}

\author{
  Phuong Huu Vu Tran\textsuperscript{1,*,**},
  Long Minh Vo\textsuperscript{2,*,**},
  Son Nguyen Minh Le\textsuperscript{1},
  Hoang Van\textsuperscript{2,3}\\
  \textsuperscript{1}Vietnamese-German University, Vietnam\\
  \textsuperscript{2}RMIT University Vietnam, Vietnam\\
  \textsuperscript{3}VANGIA INNOVATIONS, Vietnam\\
  \textsuperscript{*}These authors contributed equally.\\
  \textsuperscript{**}Corresponding authors: 10425032@student.vgu.edu.vn, s4215945@rmit.edu.vn
}

\date{}

\begin{document}
\maketitle
\begin{abstract}
We present \sys{}, the winning system for the UZH Shared Task at ArgMining 2026 on paragraph-level argument mining in UN and UNESCO resolutions. The task requires paragraph-type classification, prediction of a subset of $141$ official tags, and directed relation prediction under a strict JSON schema setting using only open-weight models up to 8B parameters. We frame the task as constrained structured prediction. The system first narrows the candidate tag space with metadata-aware dense retrieval, then applies constrained decoding with per-dimension caps, escalates only uncertain cases to a three-agent debate branch, and finally validates the output schema. On the official leaderboard, \sys{} ranked \best{1st overall}, with \best{1st in F1} and \best{5th in LLM-as-a-Judge}. During development, our configuration search further improved Task~1b Micro-F1 from $35.83\%$ to $40.08\%$ while keeping the internal Task~2 score at $4.421$. The main lesson is simple: reducing the decision space before generation improves both accuracy and submission robustness. Our code and supporting scripts are publicly available at: \href{https://github.com/LLM-Instruct-at-UZH-Shared-Task-2026/Method}{https://github.com/LLM-Instruct-at-UZH-Shared-Task-2026/Method}
\end{abstract}

\section{Introduction}
The UZH Shared Task at ArgMining 2026 asks participants to reconstruct argumentative structure in highly formal institutional texts. For each paragraph, a system must determine whether it belongs to the preambular or operative part of a document, assign a subset of 141 predefined thematic tags, and recover directed argumentative relations to other paragraphs in the same document. The task is difficult because the texts are long, the label inventory is closed and structured, and the submission format is strict: non-conforming JSON is not evaluated. In other words, the benchmark requires both semantic plausibility and schema-valid output.

This benchmark is better viewed as a constrained prediction task than as open-ended text generation. The model must reason over long institutional paragraphs, but it must also stay within a fixed label inventory, preserve paragraph indices, and produce schema-valid predictions. We therefore narrow the admissible output space before final generation. Candidate retrieval narrows the tag space; organizer-provided metadata is reused in retrieval and in per-dimension caps; decoding is projected back to the retrieved set; and final validation enforces the submission schema.

This paper makes three contributions. First, it describes the end-to-end pipeline of the first-ranked LLM-INSTRUCT submission under the official shared-task constraints. Second, it identifies the key design choice behind the result: constraining admissible tags before generation instead of asking the model to search the full 141-tag inventory. Third, it reports the development trajectory that exposed the main failure mode of early runs, namely cross-dimension over-prediction.


\section{Related Work}
Our system combines ideas from constrained decoding, dense retrieval, debate-style reasoning, and argument-mining pipelines. In particular, it is closest to settings that restrict admissible outputs before or during decoding \citep{geng2023grammar,liu2022autoregressive}, while also drawing on dense retrieval \citep{karpukhin2020dense}, debate-style control \citep{du2023improving}, and prior argument-mining work \citep{lawrence2019survey,stab2017parsing}. The difference in this shared task is that semantic plausibility alone is not enough: predictions must also satisfy a strict output schema. For broader context, our pipeline also contrasts with end-to-end and text-to-text approaches to argument mining, including structured prediction as generation \citep{paolini-etal-2021-structured}, end-to-end universal argument mining \citep{cao-2023-autoam}, fine-tuned LLM pipelines for AM \citep{cabessa-etal-2025-argument}, and recent LLM work on relation-based argument mining \citep{gorur-etal-2025-large}.

\section{Proposed Method}
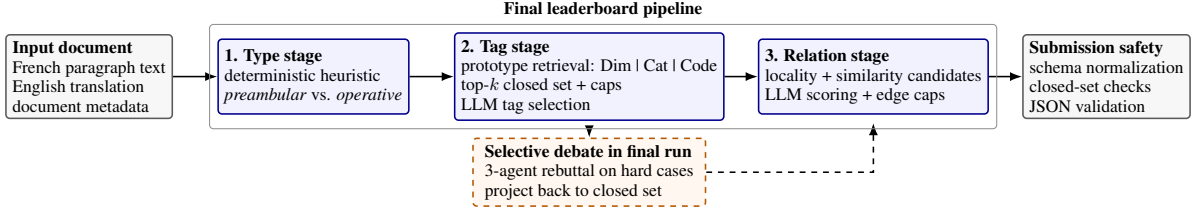
\begin{figure*}[t]
\centering
\resizebox{0.98\textwidth}{!}{%
\begin{tikzpicture}[
  font=\small,
  node distance=4mm and 5mm,
  box/.style={rounded corners=2pt, draw=black!65, thick, align=left, fill=gray!6, minimum height=8mm},
  stage/.style={rounded corners=2pt, draw=blue!55!black, thick, align=left, fill=blue!5, minimum height=10mm},
  note/.style={rounded corners=2pt, draw=black!55, thick, align=left, fill=white, minimum height=7mm},
  opt/.style={rounded corners=2pt, draw=orange!70!black, dashed, thick, align=left, fill=orange!6, minimum height=7mm},
  arrow/.style={-{Latex[length=2.2mm]}, thick}
]

\node[box, minimum width=3cm, minimum height=1.5cm] (input) {\textbf{Input document}\\French paragraph text\\English translation\\document metadata};

\node[stage, right=8mm of input, minimum width=2.7cm, minimum height=1.35cm] (type) {\textbf{1. Type stage}\\deterministic heuristic\\\emph{preambular} vs. \emph{operative}};

\node[stage, right=8mm of type, minimum width=4.2cm, minimum height=1.35cm] (tag) {\textbf{2. Tag stage}\\prototype retrieval: Dim | Cat | Code\\top-$k$ closed set + caps\\LLM tag selection}; 

\node[opt, below=3mm of tag, minimum width=4.2cm] (debate) {\textbf{Selective debate in final run}\\3-agent rebuttal on hard cases\\project back to closed set};

\node[stage, right=6mm of tag, minimum width=3.5cm, minimum height=1.35cm] (rel) {\textbf{3. Relation stage}\\locality + similarity candidates\\LLM scoring + edge caps};

\node[box, right=6mm of rel, minimum width=3.0cm, minimum height=1.35cm] (safe) {\textbf{Submission safety}\\schema normalization\\closed-set checks\\JSON validation};

\draw[arrow] (input) -- (type);
\draw[arrow] (type) -- (tag);
\draw[arrow] (tag) -- (rel);
\draw[arrow] (rel) -- (safe);
\draw[arrow,dashed] ($(tag.south)+(0,-1.5mm)$) -- (debate.north);
\draw[arrow,dashed] (debate.east) -| ($(rel.south)+(0,-1.5mm)$);

\node[draw=black!45, rounded corners=2pt, fit=(type)(tag)(rel), inner sep=4pt, label={[font=\small]above:\textbf{Final leaderboard pipeline}}] {};
\end{tikzpicture}%
}
\caption{System pipeline. The winning configuration reduces the decision space before final generation: dense retrieval creates an admissible tag set, organizer-provided metadata is encoded in the tag prototypes and reused for per-dimension caps, and high-uncertainty cases are escalated to a three-agent debate branch whose output is projected back onto the same admissible label set before final schema validation.}
\label{fig:pipeline}
\end{figure*}

Figure~\ref{fig:pipeline} summarizes the pipeline. The final leaderboard run has three prediction stages followed by a submission-safety layer. We design the pipeline to satisfy these strict task constraints while keeping generation tightly controlled. We describe its components next.

\subsection{Type stage}
Each paragraph is first classified as \textit{preambular} or \textit{operative}. In development, we found that a deterministic heuristic was more stable than a pure LLM-based alternative, so the final configuration keeps this step rule-based. Specifically, we classify paragraphs as operative when they match explicit numbered clause patterns or operative cue phrases, and otherwise default to preambular. Table~\ref{tab:type_heuristic_examples} gives representative triggers used by the heuristic. This reduces variance early in the pipeline and avoids propagating unstable type decisions to later stages.

\begin{table}[t]
\centering
\small
\caption{Representative triggers used by the type heuristic.}
\label{tab:type_heuristic_examples}
\begin{tabular}{p{0.26\columnwidth}p{0.40\columnwidth}p{0.16\columnwidth}}
\toprule
Rule type & Example opener & Predicted type \\
\midrule
French preambular cue & \textit{Considérant ...}, \textit{Reconnaissant ...}, \textit{Rappelant ...} & preambular \\
French operative cue & \textit{Demande ...}, \textit{Souligne ...}, \textit{Attire ...}, \textit{Proclame ...} & operative \\
English compound cue & \textit{Calls upon ...}, \textit{Takes note ...} & operative \\
Numbering pattern & \texttt{1.}, \texttt{2)}, \texttt{(3)} at paragraph start & operative \\
Fallback & no cue matched & preambular \\
\bottomrule
\end{tabular}
\end{table}

\subsection{Metadata-aware tag retrieval and decoding}
The tag stage is the core of the system. We read the official tag CSV and keep rows whose CODE field is neither empty nor NA. For each remaining tag $t$, let $d_t$, $c_t$, and $y_t$ denote its released dimension, category, and CODE fields. We textualize the tag prototype as
\[
p_t = d_t \,\|\, c_t \,\|\, y_t ,
\]
where $\|$ denotes string concatenation. We then embed both $p_t$ and the paragraph text with intfloat/e5-base-v2 \citep{wang2022text}, and cosine similarity is then used for dense top-$k$ retrieval \citep{wang2022text,karpukhin2020dense}.


This design makes the retrieval process metadata-aware: dimension and category information is present inside the prototype text before final tag generation. The LLM never predicts over the full 141-tag inventory. Instead, it selects from the retrieved closed set. In the final run, we apply a global cap of 5 tags per paragraph and a per-dimension cap of 2 tags. These fixed controls limit redundancy within a single semantic region and suppress cross-dimension over-prediction. Finally, any decoded label outside the retrieved candidate set is rejected. Together, these controls address the main failure mode observed during development, namely cross-dimension over-prediction.

In addition to retrieving candidate tags, the final tag prompt includes up to
three retrieved in-context examples from the training release when their
embedding similarity exceeds 0.70. These examples are used only as contextual
evidence for how similar paragraphs are tagged; they do not expand the official
tag inventory, and the final prediction is still projected back to the retrieved
closed set.

\subsection{Selective debate branch}
We also implement a hard-case route with three open-weight 8B agents-Qwen3-8B \citep{yang2025qwen3}, Llama-3.1-8B-Instruct \citep{grattafiori2024llama}, and Ministral-8B-Instruct \citep{mistral2024ministral}-inspired by debate-style reasoning \citep{du2023improving}.

Under this setup, a debate is triggered only for uncertain paragraphs, namely cases where the top-1 tag margin is low under a development-tuned routing rule, the heuristic and generator disagree on type, or the retrieved candidates show strong overlap across competing tag dimensions. Each agent proposes a type/tag hypothesis, the agents exchange one focused rebuttal round, and a constrained selector accepts only labels that remain inside the retrieved candidate set while reapplying the same per-dimension caps. Put differently, debate never expands the admissible label set; it only helps resolve hard cases within the same retrieved closed set.

\subsection{Relation stage}
Relation prediction starts from sparse candidate generation rather than exhaustive all-pairs scoring. For clarity, Table~\ref{tab:relation_labels} summarizes the four official relation labels used in Subtask~2. For each source paragraph, we form a candidate target set by taking all targets within a locality window of one paragraph and adding the top six embedding-similar targets in the same prediction instance. This keeps nearby discourse links available while still allowing non-adjacent semantic links.

The Qwen3-8B generator then scores each candidate pair using only the four official labels in Table~\ref{tab:relation_labels} or a no-edge option. We keep pairs whose confidence is at least $0.40$ and cap each source paragraph at five outgoing edges, ranked by confidence. These filters control the density of the relation graph and prevent relation prediction from becoming an unbounded all-pairs generation problem. Because the held-out test release does not expose gold relation labels, we report relation-output statistics rather than gold precision or recall for this stage.


\begin{table}[t]
\centering
\small
\caption{Official relation labels in Subtask~2.}
\label{tab:relation_labels}
\begin{tabular}{p{0.28\columnwidth}p{0.58\columnwidth}}
\toprule
Label & Informal meaning \\
\midrule
supporting & strengthens or justifies another paragraph \\
complemental & adds compatible information without conflict \\
modifying & narrows, qualifies, or conditions another paragraph \\
contradictive & conflicts with or contradicts another paragraph \\
\bottomrule
\end{tabular}
\end{table}

\subsection{Submission safety}
The final stage focuses on submission validity. It normalizes schema variants at ingestion, validates required keys before export, enforces the official relation label set, checks paragraph-index consistency, and attempts to repair malformed JSON up to three times when necessary. This component is essential because submissions that violate the required format do not receive an official score.

\section{Experiments}
\subsection{Data and Official Setting}

The shared task contains two subtasks. Subtask 1 predicts paragraph type (preambular or operative) and a multi-label subset of 141 official tags. Subtask 2 predicts paragraph-to-paragraph links and labels each directed relation as contradictive, supporting, complemental, or modifying. Official ranking averages an automated F1 metric and an LLM-as-a-Judge score. Because the public leaderboard reports ranks rather than absolute values, we report official rank positions and complement them with internal development metrics.

The organizers release a large \textit{unlabelled} training set drawn from $2,695$ UN resolutions from the UN-RES corpus associated with the SpiritRAG resource~\citep{gao2025spiritrag}, together with a held-out UNESCO evaluation split. We use the task release as provided by the organizers and do not introduce additional training labels. The texts are provided in French, with English translations provided by the task organizers to support non-French-speaking participants. In our final configuration, Task~1 and Task~2 read the English field whenever it is available and otherwise fall back to the original French paragraph. We prioritize English when available because the organizer-provided translations simplified prompt design, qualitative inspection, and debugging for a non-French-speaking team. The task page also releases a CSV file, evaluation\_dimensions\_updated.csv, containing the official tag inventory together with dimension and category metadata. Our method uses this file directly in both retrieval and decoding. In the current analyzed artifact, all evaluated instances had English available, so we do not present a separate French-only case study in this version.


\subsection{Models and Evaluation Protocol}

The final system complies with the task policy of using only open-weight models with at most 8B parameters. The main generator is Qwen3-8B in 4-bit inference mode \citep{yang2025qwen3}. Dense retrieval uses intfloat/e5-base-v2 \citep{wang2022text}. Table~\ref{tab:config} reports the deterministic settings of the best performing internal configuration.

We report two complementary views of performance. First, we present the \emph{official leaderboard results} on the test set. Because the organizers have not yet released the exact official scores, we report rank positions rather than absolute scores. Second, we report the \emph{internal development trajectory} that guided system selection and revealed the main failure mode. Following the shared-task setup, our internal evaluation uses Task~1 scores together with a Task~2 score derived from judge-based relation assessment. Code, prompts, and supporting scripts are publicly available at \href{https://github.com/LLM-Instruct-at-UZH-Shared-Task-2026/Method}{our GitHub repository}.

\begin{table}[ht]
\centering
\footnotesize
\setlength{\tabcolsep}{3.0pt}
\renewcommand{\arraystretch}{0.96}
\begin{tabular}{@{}p{0.34\columnwidth}p{0.58\columnwidth}@{}}
\toprule
\textbf{Component} & \textbf{Setting} \\
\midrule
Generator & Qwen3-8B (4-bit) \\
Reasoning budget & 256 \\
Embeddings & intfloat/e5-base-v2 \\
Language path & English if available, else French \\
Type mode & heuristic \\
Tag retrieval & cosine top-$k$, $k=40$ \\
Tag textualization & Dimensions | Categories | CODE \\
Tag decoding & threshold $0.33$; max 5 tags \\
Per-dimension cap & max 2 tags per dimension \\
Relation candidates & window $=1$, $k=6$ \\
Relation filtering & threshold $0.40$; max 5 edges/source \\
RAG examples & $k=3$, minimum cosine score $0.70$ \\
Debate & enabled for hard cases \\
JSON repair & max 3 retries \\
\bottomrule
\end{tabular}
\caption{Key settings of the strongest internal configuration and the final leaderboard submission.}
\label{tab:config}
\end{table}


\subsection{Results}


\begin{table*}[!t]
\centering
\footnotesize
\setlength{\tabcolsep}{3.0pt}
\renewcommand{\arraystretch}{0.95}
\begin{tabular}{p{0.92cm}p{5.05cm}cccccc}
\toprule
\textbf{Run} & \textbf{Dominant setting} & \textbf{T1a Acc} & \textbf{T1a F1} & \textbf{T1b P} & \textbf{T1b R} & \textbf{T1b F1} & \textbf{T2 Judge}\\
\midrule
Phase\_0 & \textit{Initial baseline from the internal script} & 72.19 & 69.69 & \best{54.92} & 26.59 & 35.83 & \best{4.421} \\
Phase\_1 & \textit{Recall-oriented tagging; looser selection increased over-prediction} & 85.81 & 83.21 & 21.57 & 30.11 & 25.13 & 4.388 \\
Phase\_2 & \textit{Further recall-oriented tuning; cross-dimension false positives remained high} & 85.77 & 83.16 & 21.53 & 30.09 & 25.10 & 4.364 \\
Phase\_3 & \textit{Final constraint-aware run with metadata-aware retrieval, selective debate, per-dimension caps, and closed-set validation} & \best{86.08} & \best{83.49} & 49.94 & \best{33.47} & \best{40.08} & \best{4.421} \\
\bottomrule
\end{tabular}
\vspace{-0.35em}
\caption{Internal development trajectory. T1a is paragraph-type classification;
T1b is tag assignment. T2 Judge is the internal LLM-as-a-Judge weighted relation
score. The main correction from Phase~1/2 to Phase~3 is precision recovery under
a similar recall regime, consistent with the over-prediction diagnosis.}
\label{tab:phase}
\vspace{-0.55em}
\end{table*}

\subsubsection{Final Rank on the Official Leaderboard}
Table~\ref{tab:official} gives the official leaderboard ranks. The main empirical outcome is straightforward: \sys{} ranked \best{1st overall}. The table also shows the strongest competing teams and our earlier submission, \textit{LLM-Instruct-2}, which corresponds to the earlier Phase2-style configuration.

\begin{table}[t]
\centering
\scriptsize
\begin{tabular}{lccc}
\toprule
\textbf{Team} & \textbf{F1} & \textbf{Judge} & \textbf{Final} \\
\midrule
\best{LLM-Instruct} & \best{1} & \best{5} & \best{1} \\
Prompteam & 5 & 1 & 2 \\
Argchestrators & 2 & 6 & 3 \\
HybridArguer & 4 & 3 & 3 \\
\midrule
LLM-Instruct-2$^{*}$ & 7 & 4 & 6 \\
\bottomrule
\end{tabular}
\caption{Official leaderboard ranks from the UZH Shared Task. $^{*}$\sys{}-2 is our first submission and corresponds to the earlier Phase2-style run.}
\label{tab:official}
\end{table}

This pattern is consistent with the design priorities of the system. The gains appear strongest when the benchmark rewards valid structured output. The official rank-1 outcome therefore supports our central claim that schema-aware control layers are useful for paragraph-level argument mining under hard output constraints. It also aligns with the internal trajectory in Table~\ref{tab:phase}, where the earlier submission \textit{LLM-Instruct-2} ranked below the final \sys{} system.

\subsubsection{Development trajectory}
Table~\ref{tab:phase} summarizes the main development phases. We explain the phases explicitly because the most useful lesson from development is not merely that the final run scored higher, but \emph{why} it scored higher.

The pattern is informative. Early recall-oriented configurations raised coarse paragraph-type scores but harmed Task~1b badly. The reason was over-prediction: loose tag selection created many cross-dimension false positives. The final run corrected this failure mode by combining metadata-aware retrieval before generation, retrieved examples in the tag prompt, per-dimension caps during selection, and strict closed-set validation after decoding.

\FloatBarrier
\subsubsection{Component diagnostics}
We next report a compact component diagnosis, keeping only the results that show clear accuracy or robustness effects. The subset ablations use the same stratified 12-document subset and the same fast decoding setting, so they should be read as relative component evidence rather than absolute final-system scores. Details are in Table~\ref{tab:component_diag}.

\begin{table}[t]
\centering
\scriptsize
\setlength{\tabcolsep}{3.2pt}
\renewcommand{\arraystretch}{0.96}
\begin{tabular}{lrrrr}
\toprule
\textbf{Variant} & \textbf{P} & \textbf{R} & \textbf{F1} & $\Delta$ \\
\midrule
Subset baseline & 37.24 & 22.76 & 28.26 & -- \\
No RAG examples & 37.52 & 16.68 & 23.09 & -5.17 \\
CODE-only prototype & 27.89 & 16.23 & 20.52 & -7.74 \\
No closed-set filter & 37.45 & 23.08 & 28.56 & +0.30 \\
\bottomrule
\end{tabular}
\vspace{-0.35em}
\caption{Subset component diagnostics for Task~1b on a stratified 12-document
subset under the same fast decoding setting. Removing retrieved examples mainly
reduces recall, while replacing metadata-aware prototypes with CODE-only
prototypes substantially reduces both precision and F1. Closed-set filtering has
little effect on F1 but is retained for schema safety.}
\label{tab:component_diag}
\vspace{-0.4em}
\end{table}

The strongest component effect comes from metadata-aware prototypes: replacing the \textit{Dimension | Category | CODE} prototype with CODE-only text reduces subset Task~1b F1 by $7.74$ points. Retrieved examples are also useful: removing them lowers recall from $22.76\%$ to $16.68\%$ and F1 by $5.17$ points. By contrast, closed-set filtering does not materially change subset F1, but it prevents out-of-candidate and invalid raw tags from reaching the final JSON. A full-corpus per-dimension-cap ablation shows a smaller regularization effect: removing the cap changes F1 from $40.26\%$ to $39.84\%$ and increases false positives from $3{,}529$ to $3{,}575$.

Box~\ref{box:qual_examples} shows a representative output.

\refstepcounter{modelbox}\label{box:qual_examples}
\begin{tcolorbox}[
title={Box~\themodelbox: Representative model output},
colback=gray!5,
colframe=black!60,
boxrule=0.3pt,
sharp corners,
fonttitle=\bfseries\footnotesize,
fontupper=\footnotesize,
left=2pt,right=2pt,top=1pt,bottom=1pt,
boxsep=0.5pt
]
\textbf{Paragraph (English).} ``While acknowledging that the number of compulsory school years may vary between countries, [the Conference] considers it desirable that the number of actual years of schooling should in no case be less than seven, and notes that this minimum is already exceeded in many countries.''\\
\textbf{Type.} operative\\
\textbf{Tags.} LAW\_REG, ISC\_1\\
\textbf{Outgoing relations.} 2, 4, 7 $\rightarrow$ complemental.\\
\textbf{Interpretation.} A legal-regulatory recommendation aligned with nearby policy paragraphs.
\end{tcolorbox}
\vspace{-0.6em}


We also inspected Phase~3 errors by tag dimension. The largest false-positive
counts came from broad dimensions such as Policy theme (623 FP), Teachers
(547 FP), Legal frameworks (388 FP), Curriculum (325 FP), and Education level
(321 FP). The most frequent false-positive tags were LAW\_REG (194 FP),
POL\_EIE (161 FP), T\_OTHER (137 FP), POL\_CUR (110 FP), and PEDAG\_OTHER
(93 FP). This supports the over-prediction diagnosis: broad policy-related
dimensions are semantically close to many paragraphs and are therefore easy to
over-select under recall-oriented prompting.

Performance also varied strongly by tag frequency. For tags appearing more than
20 times in the internal reference, Phase~3 achieved 57.22\% precision,
33.71\% recall, and 42.43\% F1. For medium-frequency tags with 6--20 references,
F1 dropped to 16.61\%; for rare tags with at most five references, F1 was only
2.51\%. Thus, the constraint-aware design reduced broad over-prediction but did
not solve sparse-label recognition.

More concretely, disabling closed-set filtering introduced 8 official tags
outside the retrieved candidate set and 5 invalid raw tags in the trace, even
though the subset F1 changed only marginally. We therefore retain this step as
a reliability guard rather than an accuracy-driven component.

\subsection{Output statistics}
\label{sec:ops}
Operationally, the final submission JSON is organized by recommendation-level TEXT\_ID values. The final artifact contains 89 prediction instances spanning 44 unique source documents, where the source document title is read from METADATA.structure.doc\_title. All runtime and output counts reported below are computed at this prediction-instance level.

Because the held-out test release does not expose gold relation labels, we cannot compute candidate recall or per-label precision/recall against gold for the final submission. We therefore report descriptive output statistics directly from the final submission artifact. Across these prediction instances, the final system produced 13{,}323 directed edges among 132{,}840 possible within-instance paragraph pairs, for a graph density of $10.03\%$. All four official relation labels are present: 8{,}949 \textit{complemental}, 4{,}199 \textit{supporting}, 160 \textit{modifying}, and 15 \textit{contradictive}. Non-adjacent links were common: 9{,}566 edges, or $71.80\%$, spanned more than one paragraph.



\subsection{Compute report}
\label{sec:compute}
The final run was executed on 2$\times$NVIDIA GeForce RTX 3090 24GB GPUs and processed the $89$ prediction instances in 1:42:03, averaging $68.81$ seconds per prediction instance, $2.07$ seconds per paragraph, or $34.5$ minutes per 1k paragraphs.

\section{Discussion}
\subsection{System Strengths and Advancements}
Four implementation choices appear central. First, metadata-aware tag prototypes are important: the CODE-only diagnostic produced the largest observed subset drop. Second, retrieved examples improve recall by supplying paragraph-level usage context without expanding the label inventory. Third, admissibility rules improve reliability by projecting predictions back to the retrieved official candidate set. Fourth, per-dimension caps act as a modest regularizer, reducing false positives on the full corpus even though their absolute F1 effect is smaller than the retrieval-related components.

\subsection{Limitations}
There are three main limitations in this work. \textbf{First}, the relation stage is less mature than the tag stage and remains sensitive to candidate generation and confidence thresholds. \textbf{Second}, our component diagnostics are not a full factorial ablation. Several
toggles are evaluated on a stratified subset under a faster decoding setting,
so they should be interpreted as relative component evidence rather than
absolute final-system scores. \textbf{Third}, the default path prioritizes English translations when available, so additional cross-lingual analysis on French-only cases would be valuable in future work.

\section{Conclusion}

We presented a compact, constraint-aware, and submission-safe system for paragraph-level argument mining in UN and UNESCO resolutions. The system combines metadata-aware dense retrieval, constraint-aware decoding, selective debate for high-uncertainty cases, sparse relation prediction, and explicit schema validation. Taken together, these results suggest that under hard output constraints, carefully designed control layers and selectively applied multi-agent reasoning can be as important as stronger generation.

\paragraph{Ethics Statement.}
Our system is intended for research benchmarking on institutional text, not autonomous legal or policy decision-making. Its design keeps intermediate decisions auditable and makes its constraints explicit.

\bibliography{references}

\end{document}